\documentclass{mva_style}
\usepackage{graphicx}
\usepackage{amsmath}
\usepackage{arydshln}

\DeclareMathOperator*{\argmax}{argmax} 

\usepackage[backend=bibtex]{biblatex}
\bibliography{bibliography.bib,handpose.bib,humanpose.bib}
\finalcopy 
\usepackage{paralist}

\begin{document}
\title{Accurate Hand Keypoint Localization on Mobile Devices} 

\author{Filippos Gouidis$^{1,2}$, Paschalis Panteleris$^1$,  Iason Oikonomidis$^1$, Antonis Argyros$^{1,2}$ \\
$^1$Institute of Computer Science, FORTH, Greece\\                 
$^2$Computer Science Department, University of Crete, Greece\\
{\tt\small \{gouidis,padeler,oikonom,argyros\}@ics.forth.gr}
}   
 
\maketitle

\section*{\centering Abstract}
\textit{
We present a novel approach for 2D hand keypoint localization from regular color input. The proposed approach relies on an appropriately designed Convolutional Neural Network (CNN) that computes a set of heatmaps, one per hand keypoint of interest. Extensive experiments with the proposed method compare it against state of the art approaches  and demonstrate its accuracy and computational performance on standard, publicly available datasets. The obtained results demonstrate that the proposed method matches or outperforms the competing methods in accuracy, but clearly outperforms them in computational efficiency, making it a suitable building block for applications that require hand keypoint estimation on mobile devices.
}

\section{Introduction}
\label{sec:intro}

-
In this work we are interested in hand 2D keypoint localization using regular RGB input. Specifically, given a conventional RGB image depicting a hand, our goal is to localize on it a set of predefined keypoints/landmarks, such as the centroid of the hand wrist and the finger joints. 
\begin{figure}[t]
\noindent
  \begin{center}
    \includegraphics[width=0.7\columnwidth]{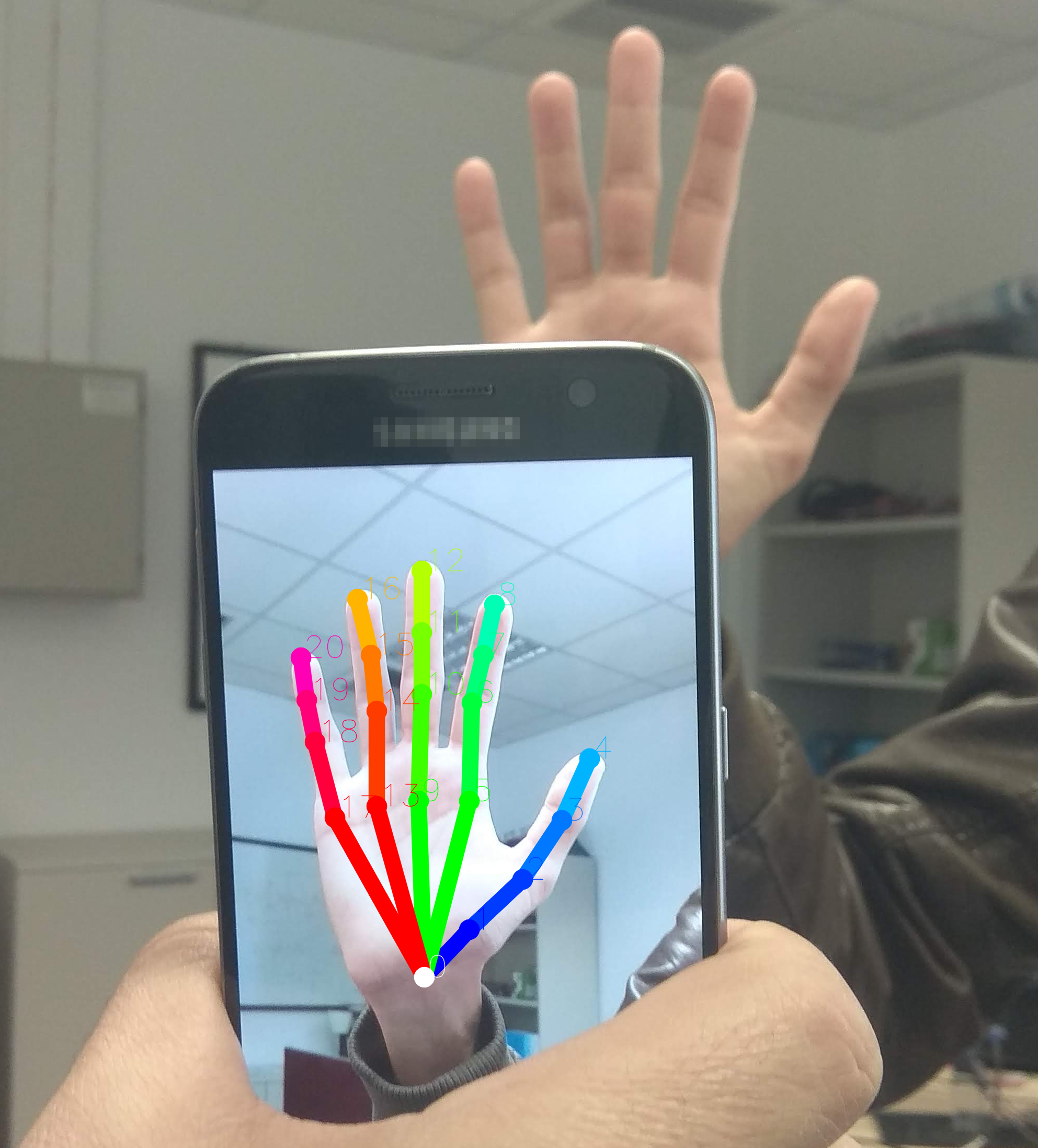}
  \end{center}
  \caption{Given an RGB image, the proposed method localizes 2D keypoints of the human hand and recovers a 2D skeletal hand model. Compared to state of the art methods, this is achieved with equivalent or better localization accuracy and with much lower computational requirements. Thus, effective hand keypoint localization becomes possible on mobile devices, enabling the development of several applications.
  }
  \label{fig:visabs}
  \vspace{-.8cm}
\end{figure}

The high interest for human hands derives from the crucial role they play in the majority of human activities. The human hands can be regarded as a general-purpose interaction tool due to their dexterous functionality in communication and manipulation. Hand landmark localization is therefore useful in many diverse cases and scenarios. As a standalone module, it can support Augmented- or Virtual-Reality applications, enabling natural interaction with the computational system. Patient assessment and rehabilitation is another area where hand keypoint localization can be immediately useful. Apart from being directly used in practical applications,
hand keypoint localization is crucial for systems that tackle more complex problems such as the estimation of the 3D pose of the observed hand. Most recent approaches on 3D hand pose estimation based on monocular RGB input incorporate a separate step~\cite{zimmermann2017learning,panteleris2017b} or at least an intermediate target of the employed neural network~\cite{iqbal2018hand,Cai2018} to the task of 2D keypoint localization.

Because of its great interest and importance in the context of numerous applications, the problem of 2D keypoints localization has attracted a lot of research interest. However, it is still considered a challenging problem that remains unsolved in its full generality. Several factors such as the hand's flexibility, self-occlusions, occlusions due to hand-object interaction, varying lighting conditions, appearance variability due to jewelry and gloves, complex/cluttered backgrounds, etc, contribute to the difficulty of the task.  
All these factors coalesce, hindering the task of accurately localizing the landmarks of interest on the observed hand.

In this work, we present an approach for 2D hand keypoint localization, based on a lightweight Convolutional Neural Network (CNN). Towards this end, we propose a modular approach, inspired by a recent successful approach on the related problem of 3D human body pose estimation from monocular input~\cite{mehta2017vnect}.
Targeting high performance even for mobile devices~\cite{ignatov2018ai}, we employ a lightweight alternative, MobileNetV2~\cite{Sandler2018}, instead of ResNet~\cite{he2016deep}, as employed by Mehta et al.~\cite{mehta2017vnect}.
Depending on the dataset, the proposed method matches or outperforms the accuracy of state-of-the-art approaches on this problem. More importantly, the proposed network achieves the fastest execution time among all competitive approaches.
Overall, it
achieves real-time, real-world performance even on mobile devices. 

In summary, our contribution can be summarized as follows:
\begin{compactitem}
    \item We design a CNN that achieves an accuracy that is similar or better to that of the state-of-the-art 2D detectors.
    \item The proposed method is based on a lightweight CNN architecture and has significantly less computational requirements than competing approaches. As such, it opens new opportunities for building hand monitoring applications on mobile devices.  
    \item We compile a new hand dataset that can be used for 2D keypoints localization. This dataset is the combination of existing, publicly available datasets. We show that training on this unified dataset improves the generalization capabilities of the resulting network.
\end{compactitem}

\section{Related Work}
\label{sec:related}
The problems of detecting and localizing human bodies and body parts, including body joints, hand joints, and human facial landmarks are of significant interest to the computer vision community and they have been actively researched for decades~\cite{moeslund2011visual,zhao2003face,Erol2007}. Considered unsolved in their full generality, they are currently actively researched. Fueled in part by the success of deep neural networks, this effort continues to yield important advancements~\cite{cao2017,jackson2017large,iqbal2018hand,Romero2017}.

\begin{figure*}[t]
\noindent
  \begin{center}
    \includegraphics[width=\textwidth]{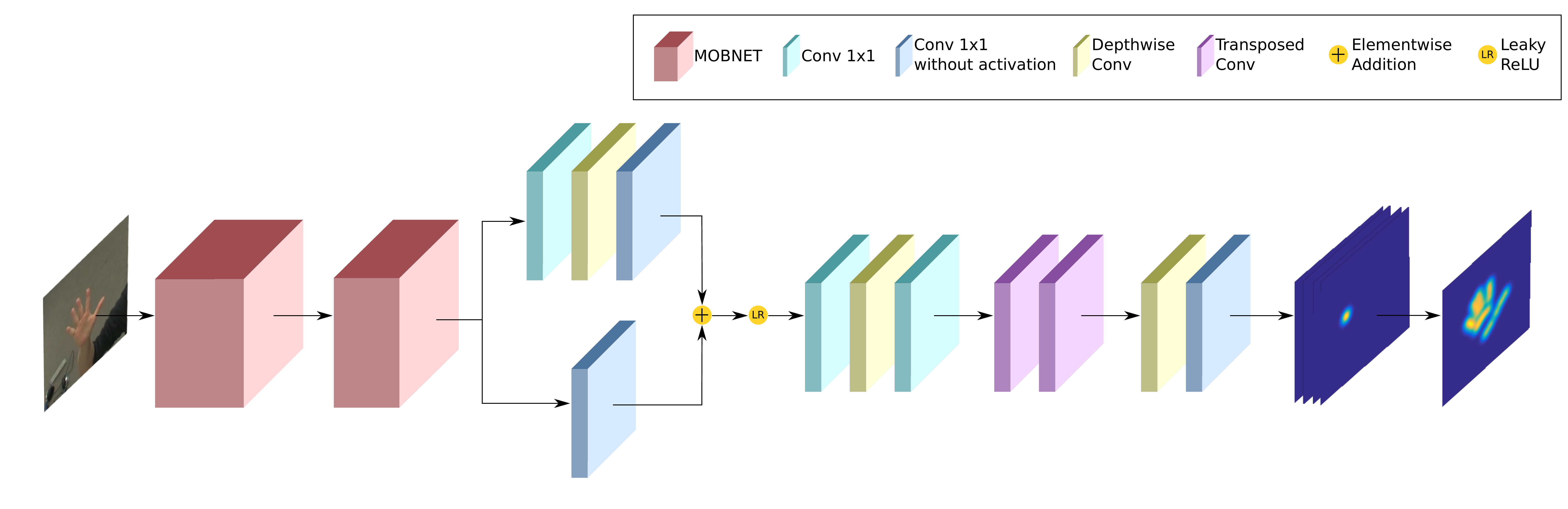} 
  \end{center}
  \vspace{-.5cm}
  \caption{
    The network architecture of the proposed approach. The input hand image (left) first passes through a modified MobileNetV2~\cite{Sandler2018} (blocks 1 to 13 and then 14 to 17 without the subsampling at the $14$th block). Next, inspired by VNect~\cite{mehta2017vnect}, a few layers compute the output heatmaps (right). See text for details.
  }
  \vspace{-.5cm}
  \label{fig:architecture}
\end{figure*}

CNNs have been successfully applied to 2D body pose estimation~\cite{wei2016convolutional,cao2017} and 2D human hand pose estimation~\cite{simon2017hand}.
This is formulated as a pixel-wise classification of each pixel being the location of a joint. As such, an important line of work in pose estimation creates networks that detect joints in 2D. Through pixel-wise classification, joint detection can exploit local patterns more explicitly than holistic regression, helping the network to learn better feature maps.

In this paper we build upon this line of work, proposing a computationally efficient 2D hand keypoint detector that matches or outperforms the accuracy of competing approaches.
Our method is closely related to the VNect work by Mehta et al.~\cite{mehta2017vnect}, borrowing and adapting the network architecture. Two more works that are closely related to our work are the one by Wei et al.~\cite{wei2016convolutional} and by Simon et al.~\cite{simon2017hand}. In contrast to all of these works, our approach targets low computational power devices, while maintaining or exceeding estimation accuracy. Further differences include the computation of 2D landmarks, in contrast to 3D ones~\cite{mehta2017vnect}, and targeting the human hand in contrast to the human body~\cite{mehta2017vnect,wei2016convolutional}. Currently, state-of-the-art methods on hand keypoint localization include the works by Simon et al.~\cite{simon2017hand}, Iqbal et al.~\cite{iqbal2018hand}, Zimmermann and Brox~\cite{zimmermann2017learning}, and Dibra et al.~\cite{dibra2018monocular}. Quantitative evaluation experiments on our method are presented in Section~\ref{sec:results}, comparing it against most of these approaches. 

During the last decade, a significant amount of research effort has been devoted to develop methods for hand pose estimation~\cite{oikonomidis2011efficient,sun2015cascaded,yuan2018}. Drawing parallels from the recent advancements on the related field of human body pose estimation~\cite{mehta2017vnect}, the effort on hand pose estimation has recently shifted from depth input, used throughout most of the past decade~\cite{yuan2018}, to regular color input~\cite{zimmermann2017learning,panteleris2017b,iqbal2018hand,Cai2018}. It should be stressed that hand keypoint localization is an important part 
of such hand pose estimation techniques.

Most recent methods on hand pose estimation can be categorized into three groups, termed {\sl generative}, {\sl discriminative}~\cite{sridhar2016real} and {\sl hybrid}. Discriminative methods~\cite{sun2015cascaded,oberweger2017deeppriorplus,iqbal2018hand}  typically learn a mapping from the visual input to the target hand pose space in a large offline step. During runtime this mapping is applied to the input, estimating the hand pose. Common ways to learn the mapping include the use of Random Forests~\cite{keskin2012hand} and more recently CNNs~\cite{oberweger2017deeppriorplus}. Our method fits in this category of approaches.
Generative methods~\cite{oikonomidis2011efficient,tzionas2016capturing,tagliasacchi2015robust} use a hand model to synthesize image features. These features are then compared to the observed visual input, quantifying the agreement between a candidate hand pose and the visual input. An optimization algorithm is then employed to search for the hand pose that best matches the visual input. Both black-box~\cite{oikonomidis2011efficient} and gradient-based~\cite{tagliasacchi2015robust} optimization algorithms have been used for this task. Finally, hybrid methods incorporate elements and techniques from both discriminative and generative ones.
\section{Localizing Hand Keypoints}
\label{sec:method}
The proposed method for hand keypoint localization is based on a convolutional neural network (CNN) that accepts as input a color image of a human hand and outputs a set of likelihood maps, or heat maps, one per landmark of interest. A post processing step is then employed to compute the final location of each landmark of interest. 
We target a total of $K=21$ hand keypoints that include the hand wrist, $5 \times 3 = 15$ finger joints, and $5$ fingertips. 

\subsection{Network Architecture}
The proposed CNN architecture is inspired by work on human body landmark localization~\cite{newell2016stacked,mehta2017vnect}. The proposed network exhibits a progressive reduction of the input spatial resolution, followed by a few steps of upsampling. This commonly used meta-architecture follows the encoder-decoder paradigm, ``encoding'' the input spatial resolution to a lower intermediate representation before ``decoding'' it by spatially upsampling.

For the encoding part, we employ a neural network that is pre-trained on the ImageNet dataset~\cite{deng2009imagenet}.
Given that we are aiming for computational efficiency, we use an adaptation of the MobileNetV2~\cite{Sandler2018}. 
More specifically, we use the structure of MobileNetV2 up to block 13 as is, without modifications.
The $14$th block is then modified by removing the stride, to disable the down-sampling that is normally performed in that block.
Then, blocks $15$ through $17$ are again used without any modifications.
For the decoding part we adapt the respective part proposed by Mehta et al.~\cite{mehta2017vnect}. In that work the target is $3D$ points, which is not the case in our work.,
Instead of this, we only output heatmaps that target 2D keypoints. Another difference is the use of less layers than their respective last part. Finally, following the example of MobileNet, we replace conventional convolutions by depth-wise convolutions, speeding up computations.
An illustration of the proposed network architecture is presented in Figure~\ref{fig:architecture}.

\subsection{Network Training}

The input of the network is regular color images while the output is $K+1$ heatmaps, one for each hand landmark, plus one for the background. This is common practice in the category of approaches using heatmaps~\cite{wei2016convolutional,cao2017}. 
We train the network on images containing only one hand.
For the $k$-th hand keypoint, $k \in \{1 \ldots K \}$, its corresponding ground truth heatmap is a 2D Gaussian, where the pixel intensity represents the likelihood of the landmark occurring in that spatial location. The Gaussian is centered at the corresponding feature point of the observed hand. 
More formally, the heatmap $H$ of the keypoint \textit{$k$} located at the 2D point $(k_x, k_y)$  has for each point $p$ in the image spatial domain $P$ with coordinates $(p_x, p_y)$ the value:
\begin{equation}
H_k(p)=e^{- \frac{(p_x-k_x)^2+(p_y-k_y)^2}{2 \sigma^2} },
\end{equation}
where $\sigma$ is a predetermined standard deviation controlling the shape of the Gaussian.
The background heatmap $H_{bg}$ is computed as the inverse mask of the per-pixel maximum over the keypoint heatmaps, that is:
\begin{equation}
    H_{bg}(p) = 1 - \max_{k \in \{ 1\ldots K \} }(H_k(p)) .
\end{equation}

\subsection{Landmark Localization}
Given an input image, the trained network computes a set of $K+1$ heatmaps.
By applying spatial softmax normalization to each of the $K$ heatmaps that correspond to landmarks (excluding the background heatmap), we obtain a probability map, where the probability at each point denotes how likely it is that the specific keypoint is located at this position. In order to obtain the position of a keypoint \textit{$k$} from a computed probability map, we select the point \textit{$\overline{k}$} having the maximum likelihood as in~\cite{wei2016convolutional, newell2016stacked, tompson2015efficient}:
\begin{equation}
\overline{k} = \argmax_{p \in P } H(p).
\end{equation}
If the probability is lower than a predetermined threshold, we select among the peaks having the highest probabilities, the one located nearest to the keypoints that have been found with high confidence.

\section{Implementation}
\label{sec:implementation}
We implemented our models using the Keras~\cite{chollet2015keras} deep learning framework. For training, we used only right hands.
During evaluation we mirror images containing a left hand~\cite{simon2017hand} before passing them to the network.

The training images are cropped around the corresponding hand center. We train for $200$ epochs, with batch size of $32$ using the ADADELTA optimizer~\cite{zeiler2012adadelta}.
We use data augmentation in order to obtain a greater variety of training samples. Namely, the images are rotated from $-30^\circ$ to $30^\circ$, translated up to $30$ pixels and scaled by a factor ranging from $0.8$ to $1.5$.
\section{Experimental Evaluation}
\label{sec:results}
We present training and evaluation details of our approach, followed by experiments that quantify the estimation accuracy and the computational performance of the proposed method. In all fronts, we compare the achieved performance against several state-of-the-art methods for 2D keypoint localization, on the basis of four different publicly available datasets.

\begin{table}[t]
\centering
\caption{\label{tab_abl} Ablation study on the CMU dataset.}
\begin{tabular}{|l|c|c|c|}
\hline
Architecture/    &  AUC $\uparrow$ &  Mean $\downarrow$ & Median  $\downarrow$ \\ 
 Training set    &       &   EPE (px)  &  EPE (px)  \\  \hline
 MobileNetv2 /   &       &             &            \\
  CMU (112x112)  & 0.917 &  11.75      & 7.34       \\  \hdashline
  CMU (224x224)  & 0.919 &  10.88      & 6.95       \\  \hdashline
  CMU+RHD        & 0.901 &  13.60      & 9.11       \\  \hdashline
CMU+RHD+SHP      & 0.924 &  10.40      & 6.40       \\  \hline
ResNet/CMU       & 0.917 &  11.14      & 6.54       \\ 
 \hline

\end{tabular}
\end{table}

\begin{figure*}[t]
\noindent
  \begin{center}
    \includegraphics[width=0.32\textwidth]{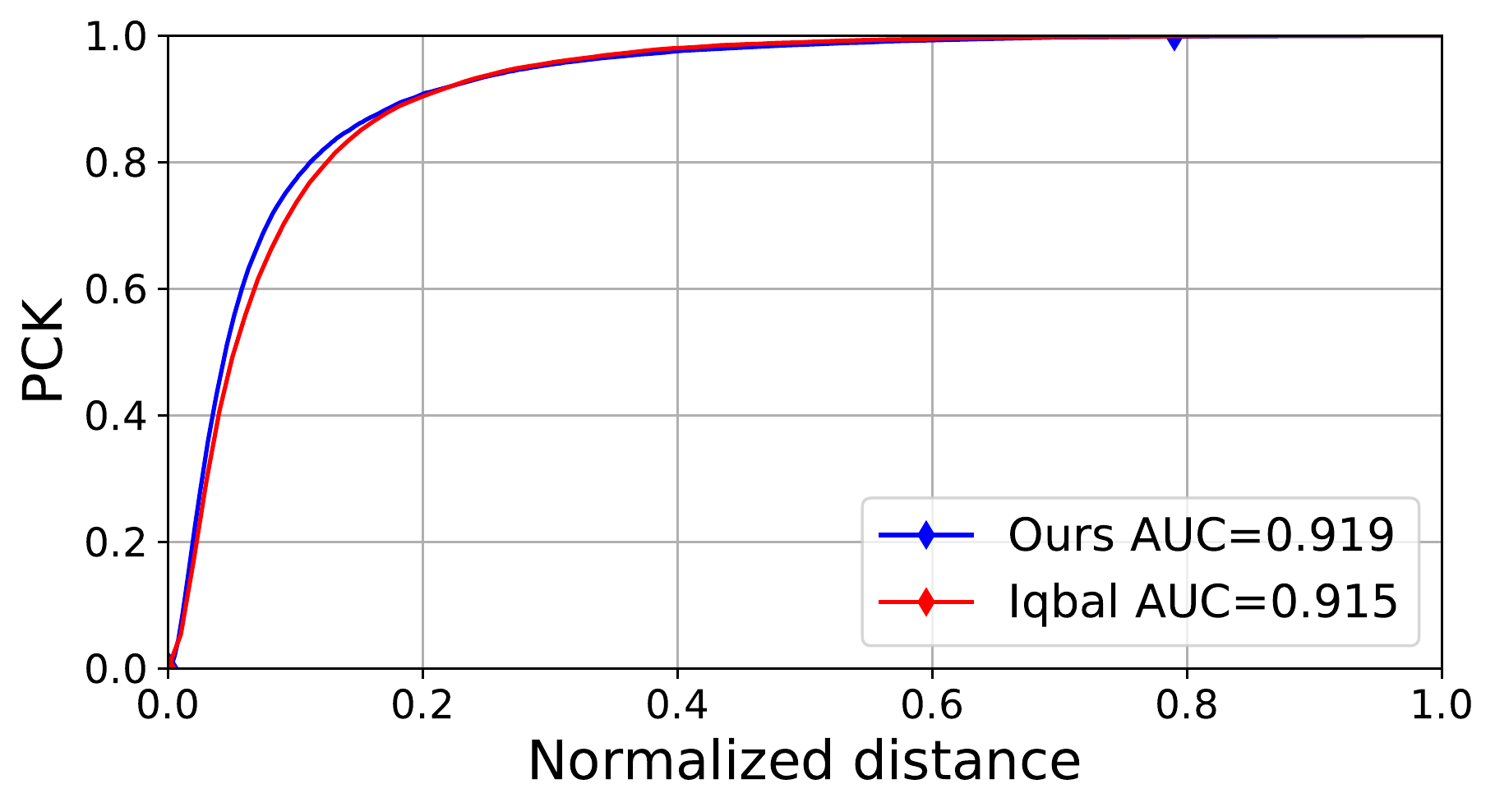}
    \includegraphics[width=0.32\textwidth]{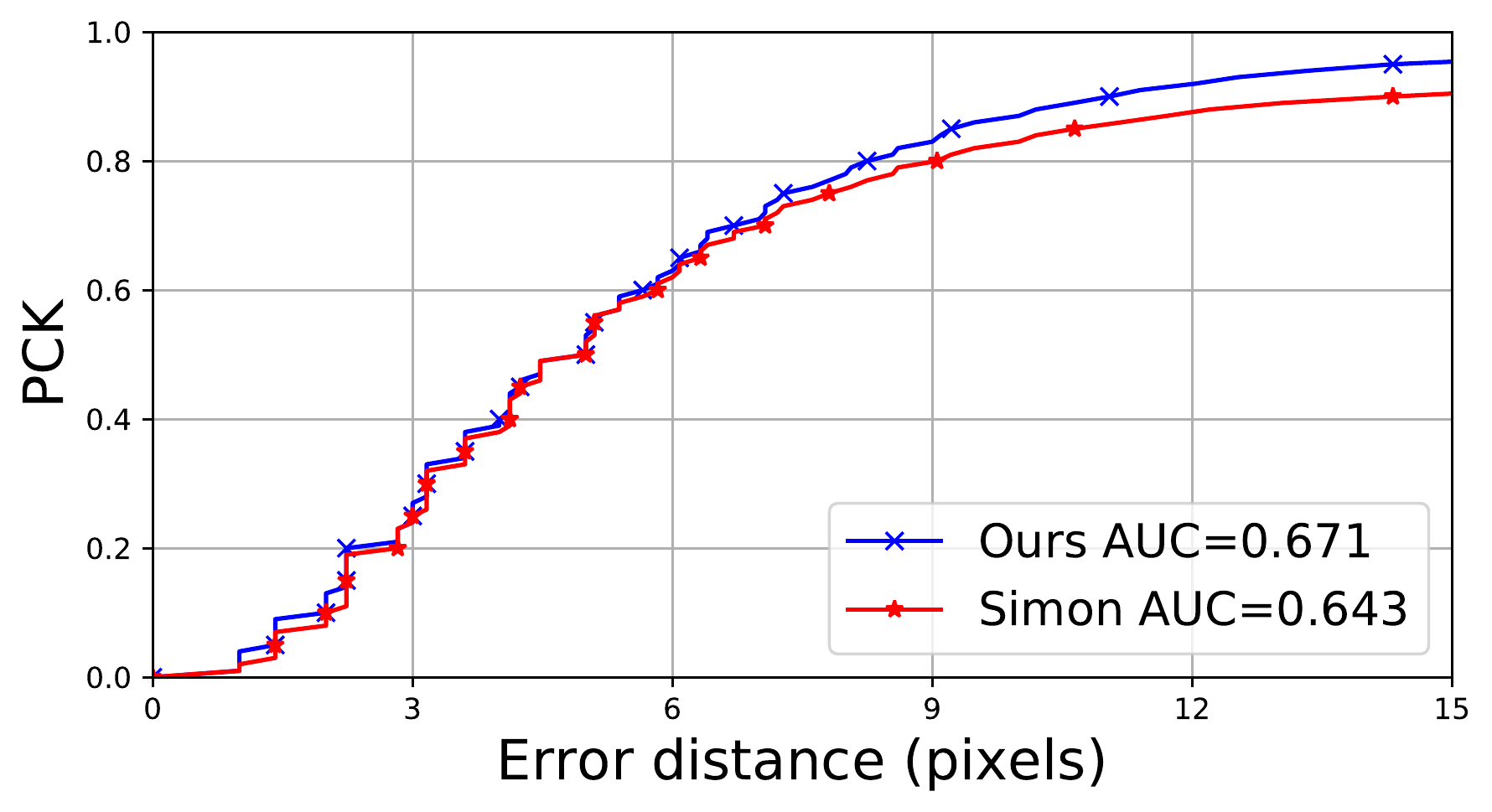}
    \includegraphics[width=0.32\textwidth]{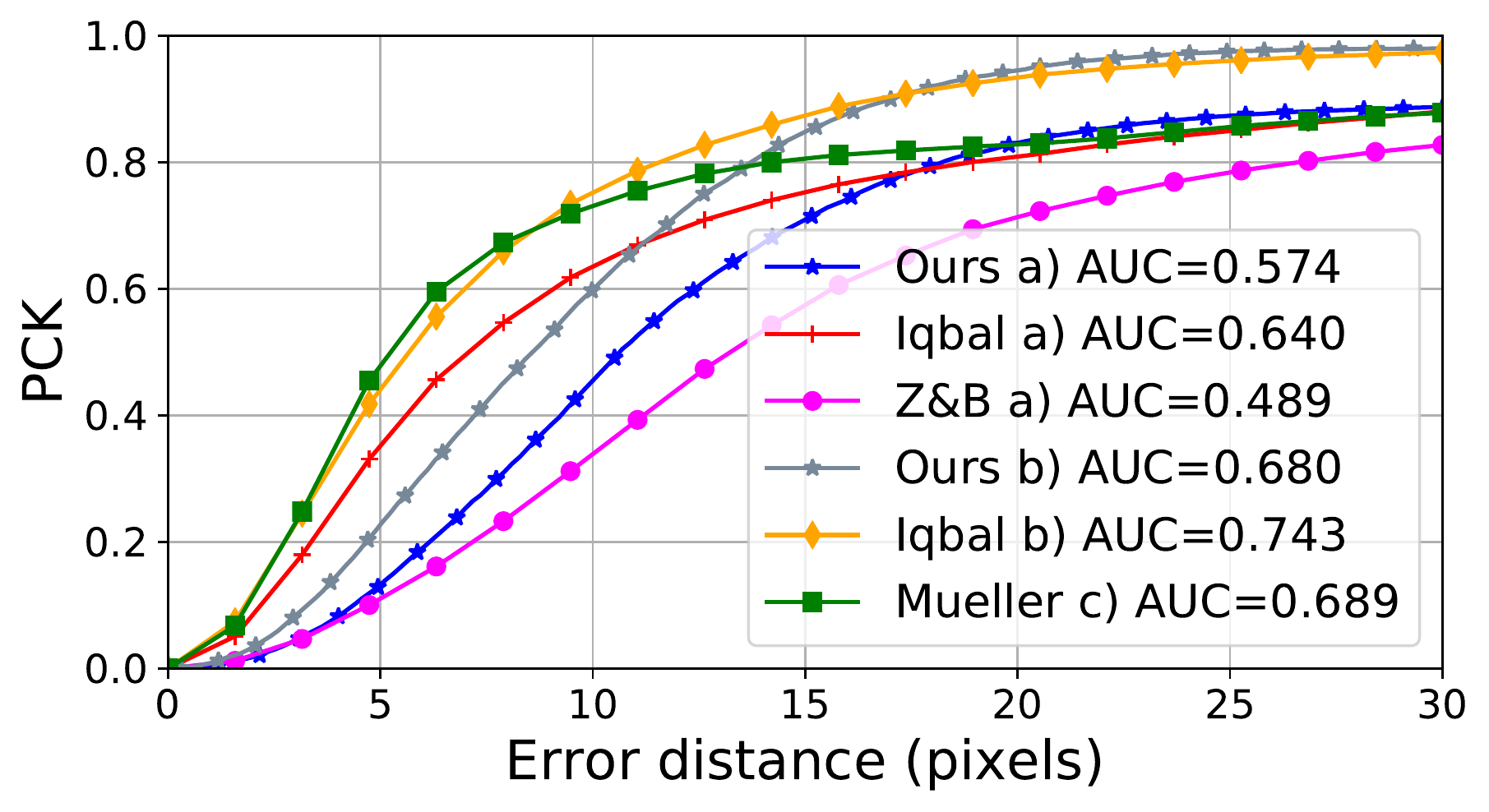}
  \end{center}
  \caption{PCK curves for the CMU (left) Tzionas (middle) and Dexter+Object (right) datasets. In the right figure a) corresponds to training on RHD and SHP dataset, b) to training on RHD, SHP and CMU dataset and c) to training on RHD, SHP and SynthHands~\cite{mueller2017real} dataset respectively.}
  \label{PCK}
\end{figure*}

\vspace*{0.3cm} {\noindent\bf Training and Test Datasets:}
We conduct experiments using the following publicly available datasets:
\begin{compactitem}
\item The \textbf{CMU dataset~\cite{simon2017hand}}: Consists of the MPII Human Pose dataset  and the New Zealand sign Language dataset (MPII + NZSL)~\cite{simon2017hand} totalling 1,912 training images and 846 images for evaluation. Additionally it also contains 14,817  real and 14,261 synthetic images that are used by the authors for their multi-view bootstrapping method.

\item The \textbf{Rendered Hand Pose (RHD) dataset~\cite{zimmermann2017learning}}: Consists of $41,258$ training and  $2,728$ evaluation synthetic images.

\item The \textbf{Stereo Hand Pose (SHP) dataset~\cite{zhang20163d}:}  Consists of $18,000$ stereo images derived from $6$ video sequences  each of which contains a single person's left hand with different background and lighting for each sequence. We use this dataset only for training.

\item The \textbf{Dexter+Object (D+O) dataset~\cite{sridhar2016real}}: Consists of $3,145$ captured frames  from $6$ video sequences each of which contains a single person manipulating an object with his left hand. We use this dataset only for testing.

\item The \textbf{Tzionas dataset \cite{tzionas2016capturing}}: Consists of $720$ captured frames from $11$ different video sequences each of which contains a person's one or two hands interacting with each other and/or with an object. We use this dataset only for testing.

\end{compactitem}

\vspace*{0.3cm} {\noindent\bf Evaluation Metrics:}
Three different metrics are commonly used throughout the relevant literature to assess quantitatively the performance of keypoint estimation methods:
(1)~The {\sl Percentage of Correct Keypoints - PCK}~\cite{iqbal2018hand} that plots the percentage of keypoints below a distance threshold as a function of this threshold, 
(2)~the {\sl Area Under Curve -  AUC} that quantifies the performance of a method by measuring the area under the PCK curve (assumes values in the range $[0..1]$, larger AUC corresponds to better performance) and 
(3)~the average {\sl End Point Error - EPE} that measures the distance in pixels between the detected joint positions and the corresponding ground truth locations.

In order to compare with state-of-the-art detectors, we adopt a common evaluation protocol for all of them. In particular, for RHD, the root keypoints of the ground-truth and estimated poses are aligned before calculating the metrics. For the CMU dataset, we report the head-normalized PCK (PCKh), similarly to~\cite{simon2017hand} and~\cite{iqbal2018hand}. For RHD and D+O  we report the PCK curve thresholded at the 30 pixels, similarly to~\cite{zimmermann2017learning} and~\cite{ iqbal2018hand}. For Tzionas, we report the PCK curve thresholded at 15 pixels.

\vspace*{0.3cm} {\noindent\bf Ablation Study:}
In order to assess the impact of different design choices upon the  performance of our solution, we conduct a number of experiments under a variety of settings.
As a target dataset in this study we opt for CMU, because we consider it the most challenging dataset due to  the wide variety of scenes it contains. The results of this study are presented in Table~\ref{tab_abl}.   
First, we examine the impact that  additional training data has on the performance of our network. We train networks with the (a)~CMU, (b)~CMU+RHD and (c)~CMH+RHD+SHP training sets. Next, we examine whether using a more complex backbone (ResNet-50) yields significantly better accuracy. Finally, we assess the impact of different crop sizes for our training images. The ablation study shows that our architecture using the Mobilenet V2 backbone trained using a crop size of $224$x$224$ yields the best trade-off for accuracy and performance.
Unless otherwise noted, 
we use this network configuration with different training sets.

\vspace*{0.3cm} {\noindent\bf Comparison to state-of-the-art:}
We test our method on the CMU, D+O, RHD and Tzionas datasets in comparison to state of the art methods that have reported results on these datasets. For a fair comparison, all methods compared on a certain dataset are trained on the same training data. Specifically, for CMU and Tzionas we use the CMU training set, whereas for the D+O and RHD datasets we use the RHD training set in combination with the SHP. 

With respect to the input to our network, we adopt the following cropping procedure. For CMU we crop the images  to a square box  centered on the hand with side size of $1.2H$, where $H$ is the size of the person's head. For RHD we crop the image to a square box centered on the hand center with side size equal to $2G$, where $G$ is the dimension of the tightest square box enclosing the hand. For D+O we use the bounding box obtained by YOLO detector~\cite{redmon2017yolo9000} which we enlarge by 25\%. For Tzionas we use a $224 \times 224$  cropped image  centered on the hand center. For the first three datasets, the cropped images are resized  to $224 \times 224$ before passed on to the network for detection.

Figure~\ref{PCK} (left) presents the comparison of our method to the methods of Iqbal et al.~\cite{iqbal2018hand} and Simon et al.~\cite{simon2017hand} on the CMU dataset. Figure~\ref{PCK} (middle) shows the comparison of our method and that of Simon et al.~\cite{simon2017hand} on the Tzionas dataset. Finally, Figure~\ref{PCK} (right) compares our method to that of \cite{simon2017hand} and \cite{zimmermann2017learning} on the D+O dataset. It should be noted that the D+O dataset has manually annotated fingertip positions which was performed on the depth frame, only. The RGBD calibration provided by the dataset authors is not accurate, so the re-projections of the fingertip positions onto the RGB frame are inaccurate.

Competitor methods compute 3D points, and they can estimate the error on the depth frame.
Our method operates on the RGB frame only, and thus its evaluation is affected by the inaccurate transfer of the ground truth on the RGB frame. This is evident in the curve of our method for the lower thresholds. For the higher thresholds where the errors due to re-projections are not dominant, our method still manages to outperform its competitors. 

On the RHD dataset the mean (median) EPE error of our method is $5.03$ $(3.11)$ pixels, whereas Iqbal et al.~\cite{iqbal2018hand} achieve a mean (median) EPE error of $3.57$ $(2.20)$ and Z\&B~\cite{zimmermann2017learning} a mean (median) error of $9.14$ $(5.00)$ pixels. We should note, however, that the aforementioned methods use an ``oracle ground truth cropping'' which, in most cases, is $4$ times smaller than the cropping bounding box that we use.

\vspace*{0.3cm} {\noindent\bf Computational Performance:}
The proposed method was designed for achieving state-of-the-art results while being able to operate on mobile devices where computational performance is limited and GPUs or NPUs are not always available. Towards that end, our MobileNet V2 backbone in combination with the depth-wise convolution blocks of the VNect-inspired module, help keep the number of parameters and FLOPs of our network very low. 
The proposed network has a total of $7.98M$ parameters and needs $16.3M$ FLOPs for processing a single frame. In comparison, the same architecture using the ResNet-50 backbone has $14.5M$ parameters and requires $29.36M$ FLOPs. 

Using TensorFlow Lite we tested our network on modern mobile phones running the Android operating system. We achieved $10$fps on the Google Pixel2 and $4$fps on the Google Pixel1 using input image size of $112 \times 112$.

On desktop hardware, our implementation achieves more than $15$fps on an i7 CPU. Furthermore, on an NVIDIA 1080TI GPU we achieve more that $200$fps ($5$ms/frame). By comparison, the work by Simon et al~\cite{simon2017hand} on the same GPU performs at $30$fps ($33$ms/frame), 7 times slower than ours.


\section{Summary}
\label{sec:conclusion}
This paper presented a novel method for accurate 2D localization of hand keypoints using regular color images. The presented method combines several ideas and tools in the relevant literature to achieve hand keypoints localization with state of the art accuracy. The computational requirements of our solution are very low, permitting real time performance on mobile devices. On-going work addresses (a)~further improvements of the presented architecture, (b)~an extension for dealing with multiple hand instances in an image, (c)~exploitation of the proposed method for 3D hand pose estimation and (d)~exploitation of the developed method in the context of HCI and HRI applications.

\printbibliography
\end{document}